\documentclass{article}
\usepackage{spconf,amsmath,graphicx,hyperref}

\usepackage{booktabs}
\usepackage{subcaption}
\usepackage{array}
\usepackage{adjustbox}
\usepackage{makecell}
\usepackage{multirow}
\usepackage{amssymb}
\usepackage{graphicx}
\usepackage{amsfonts}
\usepackage{algorithm}
\usepackage{algorithmic}
\usepackage{pifont}
\usepackage{multirow}
\usepackage{makecell}
\newcommand{\cmark}{\ding{51}}%
\newcommand{\xmark}{\ding{55}}%

\title{FastForward Pruning: Efficient LLM Pruning via Single-Step Reinforcement Learning}
\name{
\begin{tabular}{c}
Xin Yuan$^{1*}$, Siqi Li$^{1*}$, Jiateng Wei$^1$, Chengrui Zhu$^1$, Yanming Wu$^1$ \\
Qingpeng Li$^1$, Jiajun Lv$^1$, Xiaoke Lan$^3$, Jun Chen$^{2\dagger}$, Yong Liu$^{1\dagger}$
\end{tabular}
\thanks{$^*$Equal contribution.}
\thanks{$^\dagger$Corresponding authors.}
}

\address{
  $^{1}$APRIL Lab, Zhejiang University \quad $^{2}$Zhejiang Normal University \\
  $^{3}$College of Internet of Things Technology, Hangzhou Polytechnic \\
  \small\texttt{\{theoyuan,lsq4747\}@zju.edu.cn},~\texttt{junc.change@gmail.com},~\texttt{yongliu@iipc.zju.edu.cn}
}

\ninept
\begin{document}
\maketitle
\begin{abstract}
Pruning is an effective method for compressing Large Language Models (LLMs), but finding an optimal, non-uniform layer-wise sparsity allocation remains a key challenge. While heuristic methods are fast but yield suboptimal performance, more powerful search-based approaches like Reinforcement Learning (RL) are often hindered by prohibitive computational costs on large-scale models. To overcome this efficiency barrier, we propose FastForward Pruning. Its core is a decoupled, single-step RL framework that separates policy optimization from the complex budget satisfaction problem. Such a decoupling is crucial for efficiently searching the vast policy space of LLMs. This curriculum-based strategy begins with low-cost, simple tasks and gradually increases in complexity, significantly reducing the search's computational overhead. Evaluated on the LLaMA, Mistral, and OPT model families, our framework discovers pruning policies that achieve superior performance over strong heuristic baselines. Crucially, when compared to other search-based algorithms, our method achieves competitive or superior results at a fraction of the computational cost, demonstrating a clear advantage in search efficiency.

\end{abstract}

\section{Introduction}
\label{sec:intro}
Large Language Models (LLMs) have demonstrated exceptional performance across a wide range of tasks~\cite{opt, gpt3, gpt4, llama, llama2, mistral}. However, their substantial computational and storage requirements pose considerable challenges to practical deployment. Pruning provides an effective way to address this issue by reducing model size and computation. Among various pruning methodologies, structured pruning is regarded as a practical technique due to its ability to facilitate direct inference acceleration on general-purpose hardware~\cite{hoefler2021sparse}. This technique is implemented by setting a sparsity ratio for the model's prunable units, with the most straightforward approach being uniform pruning, where a single ratio is applied across all layers. However, studies indicate that layers in LLMs exhibit significant heterogeneity in pruning sensitivity~\cite{frantar2023sparsegpt, wanda, llm-pruner}. This transforms the task of pruning into a challenging optimization problem: how to determine an optimal, non-uniform sparsity allocation that maximizes model performance under a global budget.

\begin{figure}[t]
    \centering
    \includegraphics[width=\columnwidth]{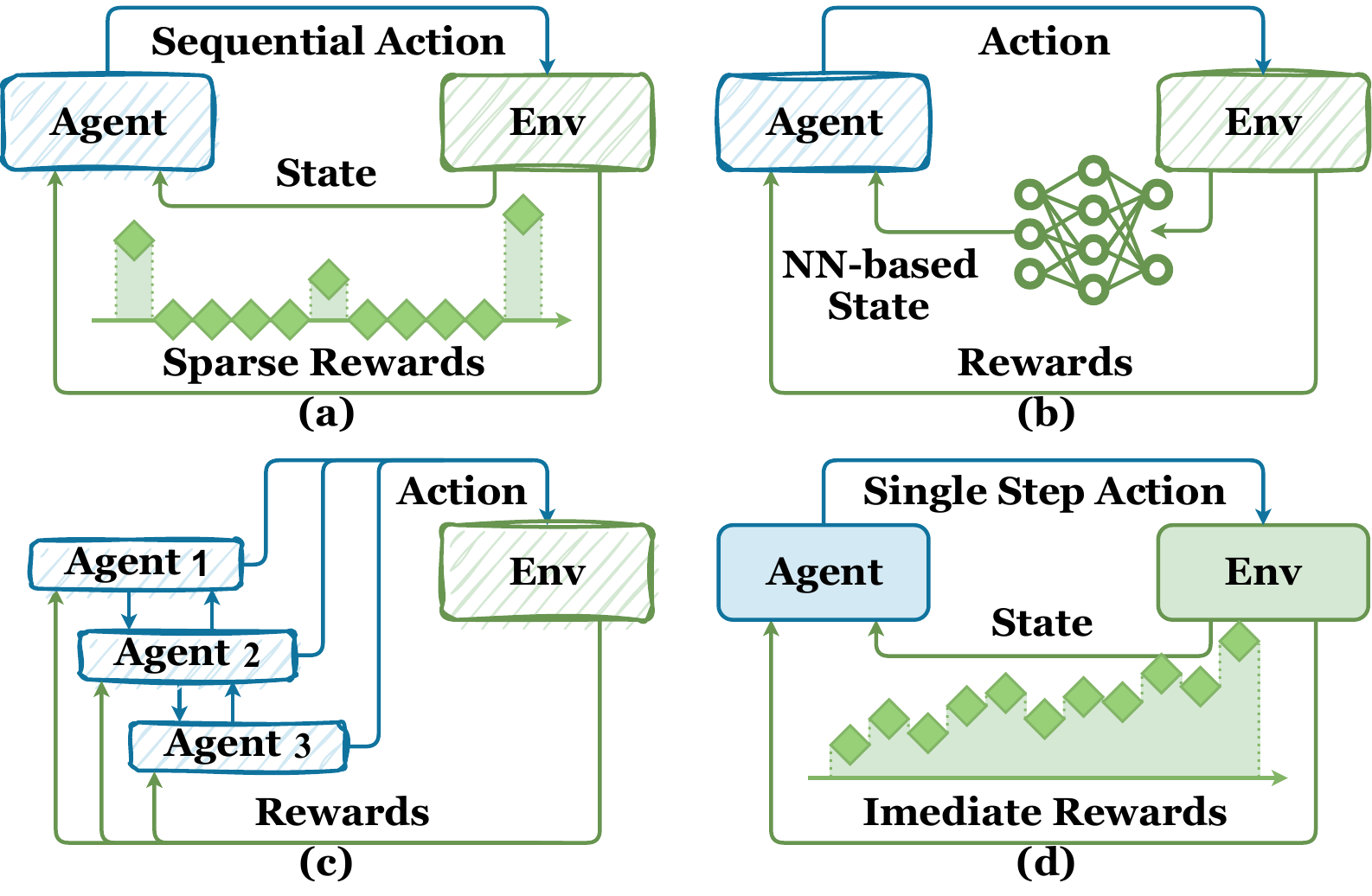}
    \caption{Our single-step RL paradigm (d) avoids the key drawbacks of legacy approaches for LLMs: (a) sparse rewards in sequential decision-making~\protect\cite{he2018amc}, (b) complex state representations~\protect\cite{yu2021autograph}, and (c) multi-agent coordination challenges~\protect\cite{alwani2022decore}.}
    \label{fig:rl_paradigms}
\end{figure}

The pursuit of this goal has primarily led to two categories of pruning methods: heuristic and search-based approaches. Heuristic strategies, such as SliceGPT~\cite{ashkboos2024slicegpt}, FLAP~\cite{an2024fluctuation}, and SVD-LLM~\cite{wang2025svdllm}, determine layer-wise sparsity or parameter reduction based on predefined, handcrafted rules. While these methods are computationally efficient and have evolved into strong performance baselines, they are often limited by their fixed, local assumptions, which may fail to capture the complex inter-layer dependencies that are crucial for preserving model capabilities~\cite{wang2020rethinking, DepGraph}. To overcome these limitations, search-based approaches were introduced to globally explore the vast policy space. These search-based methods can be divided into two main paradigms: zero-order optimization~\cite{dong2024prunerzero, shen2024search} and gradient-guided methods~\cite{n2n, he2018amc, alwani2022decore}. Zero-order methods like Evolutionary Algorithms (EAS)~\cite{dong2024prunerzero, shen2024search} suffer from immense computational cost and instability, often failing to outperform strong heuristics. Gradient-guided methods like Reinforcement Learning (RL)~\cite{PPO, SAC, DDPG} theoretically offer higher sample efficiency by using policy gradients to guide the search. However, pioneering RL frameworks are less efficient for LLMs. As illustrated in Figure~\ref{fig:rl_paradigms}, these legacy paradigms suffer from flaws that render them inefficient for large-scale models. These flaws include prohibitively slow sequential decision-making, computationally expensive state representations, and complex multi-agent coordination.

\begin{figure*}[t]
    \centering
    \includegraphics[width=0.9\linewidth]{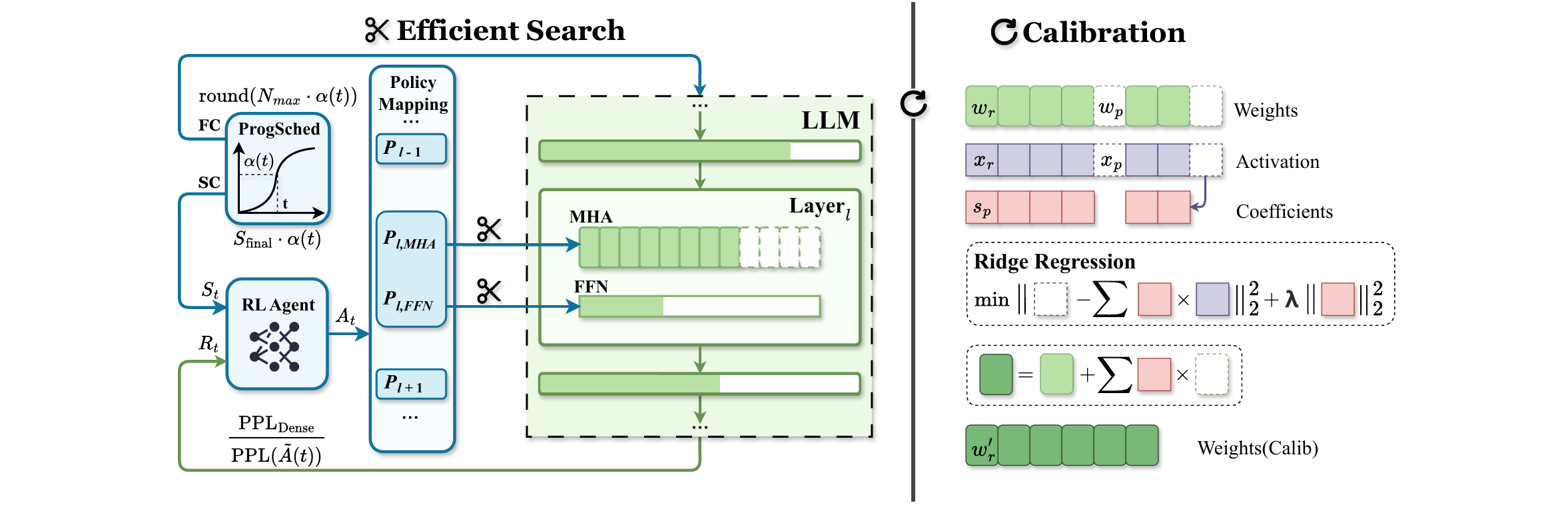} 
    \caption{An overview of our FastForward Pruning framework, consisting of two main stages. \textbf{Stage 1: Efficient Search}, where the search process is accelerated by our Progressive Scheduling (ProgSched) mechanism. It uses a unified schedule $\alpha(t)$ to govern a Sparsity Curriculum (SC) for task difficulty and a Fidelity Curriculum (FC) for evaluation cost. An RL Agent outputs raw retention scores, which are then converted into a budget-compliant policy by the Policy Mapping module. \textbf{Stage 2: Calibration}, where the final pruned weights undergo a retraining-free calibration via Ridge Regression to recover performance.
    \label{fig:overview}}
\end{figure*}

To resolve these shortcomings, we propose FastForward Pruning. Its core innovation is a decoupled design that separates the learning of layer importance from the mechanics of budget satisfaction. In this design, the RL agent learns a high-level vector of unconstrained retention scores, while a deterministic function maps these scores to a budget-compliant policy. This fundamental separation enables the formulation of the search as a stable, single-step policy generation task, avoiding the complexities of sequential RL on LLMs. To further manage the cost of this now-practical search, we introduce a Progressive Scheduling mechanism that dynamically adjusts task difficulty and evaluation fidelity. Our framework thus makes automated pruning search not just powerful, but also practical for large-scale models.

\section{Methodology}
\label{sec:methodology}

Our framework addresses the challenge of designing an efficient search strategy for discovering near-optimal, layer-wise retention policies $\mathbf{p}=[p_1, \dots, p_N]$. As illustrated in Fig.~\ref{fig:overview}, our methodology is a two-stage pipeline:  (1) an efficient search stage, where we formulate the problem as a direct policy optimization task, akin to a contextual bandit problem, which is accelerated by our novel Progressive Scheduling (ProgSched) mechanism; and (2) a performance recovery stage, which employs a lightweight, retraining-free weight calibration to compensate for pruning-induced performance loss.

\subsection{Reinforcement Learning Formulation for Pruning}
\label{ssec:rl_formulation}

A key challenge in policy learning is the credit assignment problem, which is exacerbated in a constrained action space. The global budget imposes a tight coupling among the retention rates of all layers; an increase in one layer's retention must be compensated by a decrease elsewhere. Consequently, a single policy update is a composite action, and the resulting reward signal cannot be uniquely attributed to any single dimension of the action. This ambiguity blurs the policy gradient, leading to an unstable and inefficient search.
We resolve this by decoupling policy learning from budget satisfaction. The RL agent learns a vector of unconstrained importance scores, where independent gradients provide a clear learning signal. The complex budget constraint is offloaded to a deterministic mapping function (Alg.~\ref{alg:mapping}), significantly improving search stability and sample efficiency.

\noindent\textbf{RL Environment.} We construct a focused RL environment with the sole purpose of enabling the agent to efficiently learn the relative importance of each layer.

The state ($S_t$) is defined solely by the global target sparsity ratio $\sigma_t$. This deliberate simplification allows the agent to learn a policy structure that is highly effective for transfer learning to new sparsity targets within a given model architecture, conditioned only on the global budget. This state is varied to form a learning curriculum (Sec.~\ref{ssec:acceleration}).

The action ($A_t$) is a vector of unconstrained retention scores for each layer. It represents the pure output of the learning module—the agent's raw judgment of layer importance, unpolluted by budget constraints:
\begin{equation}
\label{eq:action_space}
A_t = [a_{1}, \dots, a_{N}] \in \mathbb{R}^{N}
\end{equation}

Reward ($R_t$) evaluates the policy's final performance:
\begin{equation}
\label{eq:reward_function}
R_t = \frac{\text{PPL}_{\text{dense}}}{\text{PPL}(\tilde A_t)}.
\end{equation}
The function scales the reward signal to an appropriate range for stable gradient updates, and its scale-invariance enables the portability of our training setup across models of different scales. To optimize for this reward, we employ Proximal Policy Optimization (PPO)~\cite{PPO}, a robust and sample-efficient standard for policy optimization in high-dimensional continuous action spaces.

\noindent\textbf{Budget-Constrained Policy Mapping.} The agent's learning loop operates solely on the ($S_t$, $A_t$, $R_t$) tuple. After it outputs the unconstrained action $A_t$, all subsequent procedures are encapsulated within a deterministic execution mechanism, shielding the learning algorithm from their internal complexity.
\begin{algorithm}[t]
    \caption{Budget-Constrained Policy Mapping $\mathcal{F}$}
    \label{alg:mapping}
    \textbf{Input:} Raw action $A_t$, total preservation ratio $P$, parameters per unit $\mathbf{w}=[w_1, \dots, w_N]$, bounds $[a_{min}, a_{max}]$. \\
    \textbf{Output:} Valid action $\tilde{A}_t=[\tilde{a}_1, \dots, \tilde{a}_N]$.
    
    \begin{algorithmic}[1]
    \STATE $A'_t \leftarrow \text{clip}(\tanh(A_t+1)/2, 0, 1)$
    \STATE $A''_t \leftarrow A'_t * (a_{max}-a_{min}) + a_{min}$

    \STATE $\tilde{A}_t \leftarrow [a_{min}, \dots, a_{min}]$ \COMMENT{Initialize with lower bound}
    \STATE $B_{rem} \leftarrow (P \cdot \sum w_i) - \sum \tilde{a}_i w_i$ \COMMENT{Calculate remaining budget}
    \STATE $I \leftarrow \text{argsort}(A''_t, \text{descending})$ \COMMENT{Get indices sorted by preference}
    \FOR{$i$ in $I$}
        \STATE $\Delta_{max} \leftarrow (a_{max} - \tilde{a}_i) \cdot w_i$
        \STATE $\Delta \leftarrow \min(B_{rem}, \Delta_{max})$
        \STATE $\tilde{a}_i \leftarrow \tilde{a}_i + \Delta / w_i$
        \STATE $B_{rem} \leftarrow B_{rem} - \Delta$
        \STATE \textbf{if} $B_{rem} \le 0$ \textbf{then break}
    \ENDFOR
    
    \STATE $\tilde{A}_t \leftarrow \text{DiscretizeAndCorrect}(\tilde{A}_t, P, \mathbf{w})$
    \end{algorithmic}
\end{algorithm}

The deterministic execution mechanism first maps the agent's action $A_t$ to a budget-compliant policy $\tilde{A}_t$ via function $\mathcal{F}$ (Alg.~\ref{alg:mapping}). The final DiscretizeAndCorrect step ensures budget precision; it first rounds all retention rates to a fixed precision (e.g., 0.01), then greedily adjusts the rate of the layer with the highest initial importance score ($A''_t$) to precisely meet the budget, resolving any residual discrepancy. The actual pruning is then performed based on this policy. While we choose Wanda~\cite{wanda} as the underlying criterion for its efficiency, our framework's focus on a high-level importance policy suggests potential compatibility with other criteria, making the exploration of its synergy a promising future direction.

\subsection{Search Acceleration via Progressive Scheduling}
\label{ssec:acceleration}

Our Progressive Scheduling mechanism guides the search agent along an optimized path, from easy-to-solve problems to the complex final target~\cite{bengio2009curriculum}. This curriculum-based approach simultaneously addresses two primary efficiency bottlenecks: (1) high evaluation cost, as each candidate policy requires a computationally expensive evaluation on the LLM; and (2) invalid reward signals, as initial random policies at high sparsity can cause model collapse resulting in non-finite perplexity values, which wastes computational cycles.

To address both issues in concert, we design a Progressive Scheduling mechanism. Its core is a unified progression schedule, $\alpha(t)$, governed by a Sigmoid function:
\begin{equation}
\label{eq:alpha_curriculum}
\alpha(t) = \alpha_{start} + \frac{1 - \alpha_{start}}{1 + e^{-k(t_0 - t)}}
\end{equation}
where $t$ is the current training step, while $k$ and $t_0$ control the schedule's steepness and midpoint, respectively.
This unified schedule simultaneously governs a Sparsity Curriculum (SC) for task difficulty and a Fidelity Curriculum (FC) for evaluation fidelity, resolving the aforementioned bottlenecks.

The SC creates an easy-to-hard task curriculum by progressively increasing the target sparsity, which resolves the issue of learning stagnation from invalid initial policies:

\begin{equation} \sigma_t = S_{\text{final}} \cdot \alpha(t) \end{equation}

Meanwhile, the FC implements an adaptive evaluation fidelity strategy by adjusting the number of samples for reward evaluation (perplexity calculation), which addresses the high evaluation cost:

\begin{equation} N_{\text{eval\_samples}}(t) = \text{round}(N_{\text{max}} \cdot \alpha(t)) \end{equation}

This unified approach simplifies hyperparameter tuning and is found to be robust in our experiments.

The easy-to-hard principle embodied in our task curriculum can be further extended to search tasks across multiple sparsity targets. To this end, we employ a policy transfer mechanism: agent weights fully trained on a lower-sparsity task are used to initialize a higher-sparsity task, substantially reducing exploration time.

\subsection{Efficient Post-Pruning Calibration}
\label{ssec:calibration}
After our search strategy discovers the final layer-wise retention policy $\mathbf{p}$, we employ a computationally manageable weight calibration step to fine-tune the weights of the remaining channels, further enhancing model performance without retraining.
We adapt an established technique from prior work~\cite{udfc} to compensate for information loss.
This is formulated as a Ridge Regression problem:
\begin{equation}
\label{eq:ridge_objective}
\min_{\mathbf{s}_p} \left\| X_{p} - X_{\mathcal{R}} \mathbf{s}_p \right\|_{2}^{2} + \lambda \| \mathbf{s}_p \|_{2}^{2}
\end{equation}
where $X_p$ is the activation vector of a pruned channel, $X_{\mathcal{R}}$ is the matrix of activations from retained channels, and $\mathbf{s}_p$ are the calibration coefficients. The optimal coefficient matrix $\mathbf{S}^*$, where each column $\mathbf{s}_p^*$ represents the coefficients to reconstruct a pruned channel $p$ from the retained channels $\mathcal{R}$, is computed efficiently. These coefficients are then used to update the weights of the retained channels:
\begin{equation}
\label{eq:weight_update}
W'_{\mathcal{R}} \leftarrow W_{\mathcal{R}} + \mathbf{S}^{*T} W_{\mathcal{P}}
\end{equation}
where $W_{\mathcal{R}}$ and $W_{\mathcal{P}}$ are weight matrices corresponding to retained and pruned channels.

\section{Experiments}
\label{sec:experiments}

\begin{table*}[t!]
\centering
\caption{Zero-shot performance of structured pruning on LLaMA-V1-7B at 20\% sparsity. The \underline{underlined} results indicate the second-best performance. We report the average accuracy over seven datasets: BoolQ~\cite{boolq}, PIQA~\cite{piqa}, HellaSwag~\cite{hellaswag}, WinoGrande~\cite{sakaguchi2021winogrande}, ARC-e/c~\cite{arc}, and OBQA~\cite{openbook}.
}
\label{zs}
\resizebox{\linewidth}{!}{%
\begin{tabular}{lcclcccccccc}
\toprule
\textbf{Method} & \textbf{\makecell{Searched \\ Sparsity}} & \textbf{\makecell{Weight \\ Calib.}} & \textbf{\makecell{Calib \\ Samples}} & \textbf{BoolQ} & \textbf{PIQA} & \textbf{HellaSwag} & \textbf{WinoGrande} & \textbf{ARC-e} & \textbf{ARC-c} & \textbf{OBQA} & \textbf{Ave.} \\ 
\midrule
Dense & - & - & \multicolumn{1}{c}{-} & 75.02 & 79.16 & 76.20 & 70.09 & 72.85 & 44.62 & 44.40 & 66.05 \\ 
\midrule
SliceGPT & \xmark & \cmark & 1024 $\times$ 2048 & 58.99 & 69.86 & 59.45 & \textbf{68.43} & 62.37 & 36.60 & 37.40 & 56.15 \\
FLAP & \xmark & \cmark & 1024 $\times$ 128 & 68.59 & 74.21 & 64.98 & 64.40 & 59.89 &  \textbf{37.80} & \textbf{40.20} & 58.58 \\
SVD-LLM & \cmark & \cmark & 256 $\times$ 2048 & 64.92 & 69.64 & 60.68 & 64.96 & 56.27 & 34.56 & 39.40 & 55.78 \\ 
EAS-based & \cmark & \cmark & 128 $\times$ 2048 & 70.98 & 74.92 & 67.29 & 64.64 & 64.23 & 36.52 & 39.40 & 59.71 \\ 
\midrule
Ours(C4) & \cmark & \cmark & 32 $\times$ 2048 & \underline{72.75} & \textbf{75.73} & \textbf{69.34} & \underline{68.27} & 66.53 & 36.59 & 39.14 & \underline{61.19} \\
Ours(Calib) & \cmark & \cmark & 32 $\times$ 2048 & \textbf{74.37} & \underline{75.29} & \underline{68.41} & 67.85 & \textbf{70.23} & \underline{37.47} & \underline{39.60} & \textbf{61.89} \\ 
\bottomrule
\end{tabular}%
}
\end{table*}

\subsection{Experiment Setup}
\label{ssec:setup}

\noindent\textbf{Models and Baselines.} We evaluate our framework on OPT~\cite{opt}, LLaMA-V1~\cite{llama}, and LLaMA-V2~\cite{llama2} models against SliceGPT~\cite{ashkboos2024slicegpt} and FLAP~\cite{an2024fluctuation}, and the EAS-based search method~\cite{shen2024search}.

\noindent\textbf{Evaluation Metrics.} Performance is measured by perplexity (PPL) on the WikiText-2 dataset~\cite{wikitext} and the average accuracy across seven zero-shot common-sense reasoning tasks, evaluated via LM-Evaluation-Harness~\cite{eval-harness}. For Ours(Calib), the calibration set consists of 32 samples from the WikiText dataset, which is disjoint from the downstream evaluation tasks.

\noindent\textbf{Implementation Details.} Our policy network is a two-layer MLP (256-128 hidden units with ReLU activation) optimized via PPO (lr=1e-4, clip=0.2). The Progressive Scheduling mechanism (Eq.~\ref{eq:alpha_curriculum}) is configured with $k=0.01$ and $t_0=500$. The Ridge Regression coefficient $\lambda$ (Eq.~\ref{eq:ridge_objective}) is set to 0.01. The `DiscretizeAndCorrect` function in Alg.~\ref{alg:mapping} performs final rounding to meet the budget. Experiments are conducted on NVIDIA RTX 4090 and A800 GPUs. Our code will be publicly released.

\subsection{Main Performance}
\label{ssec:main_performance}
As shown in Table~\ref{zs}, on the zero-shot tasks, our primary method, Ours(Calib), achieves an average accuracy of 61.89, outperforming both FLAP (58.58) and the EAS-based method (59.71). This suggests our RL agent effectively preserves structures crucial for reasoning. The Ours(C4) variant, calibrated on the C4 dataset~\cite{c4}, affirms the robustness of our method by achieving strong competitive results even when the calibration data distribution differs from the test set. Furthermore, as presented in Table~\ref{tab:ppl_results_combined}, the PPL of Ours(Calib) is highly competitive across all models and sparsity levels, validating our ability to efficiently find high-quality policies.

\begin{table}[h!]
\centering
\caption{Comprehensive Perplexity (PPL, $\downarrow$) on WikiText-2 across a wide range of OPT, LLaMA-V1, and LLaMA-V2 models. The dash ‘-’ represents results that cannot be reproduced with the open-source code.}
\label{tab:ppl_results_combined}
\setlength{\tabcolsep}{4pt}
\resizebox{\columnwidth}{!}{%
\begin{tabular}{l l ccc cc cc}
\toprule
& & \multicolumn{3}{c}{\textbf{OPT}} & \multicolumn{2}{c}{\textbf{LLaMA-V1}} & \multicolumn{2}{c}{\textbf{LLaMA-V2}} \\
\cmidrule(lr){3-5} \cmidrule(lr){6-7} \cmidrule(lr){8-9}
\textbf{Sparsity} & \textbf{Method} & \textbf{125M} & \textbf{1.3B} & \textbf{2.7B} & \textbf{7B} & \textbf{13B} & \textbf{7B} & \textbf{70B} \\
\midrule
0\% & Dense & 27.64 & 14.61 & 12.46 & 5.68 & 5.09 & 5.47 & 3.32 \\
\midrule
\multirow{6}{*}{20\%} 
& SliceGPT & 34.10 & 16.58 & 13.89 & 6.99 & 6.13 & 6.86 & 4.25 \\
& FLAP & 34.45 & 17.37 & 15.38 & 6.90 & 6.05 & 7.15 & 4.12 \\
& SVD-LLM & 38.86 & 17.82 & 15.22 & 7.89 & - & 8.38 & 4.66 \\
& EAS-based & 44.12 & 19.23 & 16.44 & 6.89 & 5.90 & - & - \\
\cmidrule(l){2-9}
& Ours & 31.44 & 16.75 & 14.55 & 7.25 & 6.22 & 7.54 & 4.32 \\
& Ours (Calib) & \textbf{30.67} & \textbf{15.82} & \textbf{13.75} & \textbf{6.64} & \textbf{5.67} & \textbf{6.79} & \textbf{4.06} \\
\midrule
\multirow{6}{*}{30\%} 
& SliceGPT & 44.23 & 19.60 & 16.31 & 8.69 & 7.35 & 8.64 & 5.41 \\
& FLAP & 40.05 & 20.77 & 18.31 & 8.23 & 6.97 & 8.85 & 4.82 \\
& SVD-LLM & 49.13 & 20.68 & 17.79 & 9.52 & - & 10.66 & 5.44 \\
& EAS-based & 80.84 & 26.82 & 23.48 & 8.28 & 6.67 & - & - \\
\cmidrule(l){2-9}
& Ours  & 39.53 & 21.35 & 17.51 & 8.65 & 7.06 & 8.73 & 4.88 \\
& Ours (Calib) & \textbf{36.19} & \textbf{18.65} & \textbf{16.23} & \textbf{8.02} & \textbf{6.43} & \textbf{8.08} & \textbf{4.62} \\
\bottomrule
\end{tabular}%
}
\end{table}

\subsection{Analysis of Search Efficacy}
\label{ssec:analysis}
\noindent\textbf{Ablation Study.} The results in Table~\ref{tab:ablation_full} reveal a crucial interplay between our search and calibration stages. While the searched policy alone shows a higher initial perplexity, the final framework (Search + Calib.) significantly outperforms uniform pruning with calibration (Calib.), e.g., 6.79 vs. 7.15 PPL on LLaMA-V2 7B. This suggests our RL agent excels not at minimizing immediate disruption, but at finding a non-uniform structure whose error is more predictable and thus highly amenable to linear compensation. We also find our curriculum to be crucial, as it prevents search failure at high sparsities and noticeably accelerates convergence. The search effectively optimizes for the post-calibration potential, a key factor for the framework's success.

\noindent\textbf{Comparative Performance Analysis.} Our framework's effectiveness can be analyzed from two perspectives. When compared to heuristic methods, our search-based approach consistently yields superior performance. For instance, on LLaMA-V1 7B at 20\% sparsity, Ours (Calib) significantly outperforms FLAP in both PPL (6.64 vs. 6.90) and average zero-shot accuracy (61.89 vs. 58.58). This justifies the one-time search cost in fidelity-critical scenarios where performance is paramount.
When compared to other search-based methods like EAS, our framework's core advantage in efficiency is clear. While zero-order methods like EAS rely on brute-force exploration, our gradient-guided RL approach employs a more sophisticated search strategy. As shown in Table~\ref{tab:search_cost_perf_v1}, this translates to a direct practical advantage: our method is not only 3.4x faster but also discovers a better-performing policy (6.64 vs. 6.89 PPL). Policy transfer further enables finding the 30\% policy with only 0.9 additional hours, validating the strategy's efficiency across multiple targets. The learned policies consistently preserve middle-layer FFNs and early attention blocks, confirming their known structural importance.

\begin{table}[h!]
\centering
\caption{Ablation study on WikiText PPL ($\downarrow$) at 20\% sparsity.}
\label{tab:ablation_full}
\setlength{\tabcolsep}{4pt}
\resizebox{\columnwidth}{!}{%
\begin{tabular}{l cc ccc cc}
\toprule
\multirow{2}{*}{\textbf{Method}} & \multirow{2}{*}{\textbf{\makecell{Searched \\ Sparsity}}} & \multirow{2}{*}{\textbf{\makecell{Weight \\ Calib.}}} & \multicolumn{3}{c}{\textbf{OPT}} & \multirow{2}{*}{\textbf{\makecell{LLaMA-V2 \\ 7B}}} & \multirow{2}{*}{\textbf{\makecell{Mistral \\ 7B}}} \\
\cmidrule(lr){4-6}
& & & \textbf{125M} & \textbf{1.3B} & \textbf{2.7B} & & \\
\midrule
Dense & - & - & 27.64 & 14.61 & 12.46 & 5.47 & 5.25 \\
\midrule
Calib. & \xmark & \cmark & 32.25 & 17.15 & 14.65 & 7.15 & 7.22 \\
Search & \cmark & \xmark & 31.44 & 16.75 & 14.55 & 7.54 & 6.89 \\
Search + Calib. & \cmark & \cmark & \textbf{30.67} & \textbf{15.82} & \textbf{13.75} & \textbf{6.79} & \textbf{6.48} \\
\bottomrule
\end{tabular}%
}
\end{table}

\begin{table}[h!]
\centering
\caption{Search cost and performance comparison on LLaMA-V1 7B. Total cost is the sum of search and calibration time. $*$ Indicates the incremental search cost, as the 30\% search is warm-started from the 20\% policy.}
\label{tab:search_cost_perf_v1}
\setlength{\tabcolsep}{4pt}
\resizebox{\columnwidth}{!}{%
\begin{tabular}{l ccc ccc}
\toprule
\multirow{2}{*}{\textbf{Method}} & \multicolumn{3}{c}{\textbf{20\% Sparsity}} & \multicolumn{3}{c}{\textbf{30\% Sparsity}} \\
\cmidrule(lr){2-4} \cmidrule(lr){5-7}
& \textbf{\makecell{Search Cost \\ (GPU-hr)}} & \textbf{\makecell{Total Cost \\ (GPU-hr)}} & \textbf{\makecell{PPL \\ $\downarrow$}} & \textbf{\makecell{Search Cost \\ (GPU-hr)}} & \textbf{\makecell{Total Cost \\ (GPU-hr)}} & \textbf{\makecell{PPL \\ $\downarrow$}} \\
\midrule
FLAP & - & 0.02 & 6.90 & - & 0.02 & 8.23 \\
SliceGPT & - & 0.43 & 6.99 & - & 0.42 & 8.69 \\
EAS-based & 23.6 & 23.8 & 6.89 & 24.4 & 24.6 & 8.28 \\
\midrule
Ours (Calib) & 6.13 & 7.10 & \textbf{6.64} & $\ \,$7.03${}^*$ & 8.12 & \textbf{8.02} \\
\bottomrule
\end{tabular}%
}
\end{table}

\section{Conclusion}
\label{sec:conclusion}
We propose FastForward Pruning, an efficient reinforcement learning framework that solves the efficiency bottlenecks in automated LLM pruning by integrating a decoupled, single-step RL environment, a novel Progressive Scheduling mechanism, and a lightweight weight calibration module. Experiments demonstrate our method discovers pruning policies that surpass strong heuristics and rival computationally intensive search approaches, all at a fraction of the computational cost. Our work bridges the gap between fast but suboptimal heuristics and powerful but costly search methods. By achieving competitive performance at a fraction of the computational cost, we demonstrate that efficient, automated search is not just a theoretical possibility but a practical and accessible path to high-performance LLM compression.

\vfill
\pagebreak

\section{acknowledgements}
This work was supported by the Key R\&D Project of Zhejiang Province under Grant 2025C01090

\bibliographystyle{IEEEbib}

\bibliography{refs}

\end{document}